\documentclass[runningheads]{llncs}

\usepackage{iciap}

\usepackage{iciapabbrv}
\usepackage{graphicx}
\usepackage{booktabs}
\usepackage{adjustbox}
\usepackage{enumitem}

\usepackage[accsupp]{axessibility}  

\usepackage[pagebackref,breaklinks,colorlinks,citecolor=iciapblue]{hyperref}

\usepackage{orcidlink}
\usepackage{comment}
\usepackage{graphicx}
\usepackage{pifont}
\usepackage{multirow}
\usepackage{amssymb}
\usepackage{amsmath} 

\usepackage{pifont}
\newcommand{\cmark}{\ding{51}}%
\newcommand{\xmark}{\ding{55}}%

\definecolor{orange}{RGB}{255,229,204} 
\definecolor{lred}{RGB}{255,204,204} 
\definecolor{lgreen}{RGB}{200,240,200} 
\definecolor{lblue}{RGB}{204,229,255} 

\definecolor{darkbrown}{rgb}{0.4, 0.26, 0.13}
\definecolor{amethyst}{rgb}{0.6, 0.4, 0.8}
\definecolor{blue-violet}{rgb}{0.54, 0.17, 0.89}
\definecolor{caputmortuum}{rgb}{0.35, 0.15, 0.13}
\definecolor{darkviolet}{rgb}{0.58, 0.0, 0.83}
\definecolor{lavender}{rgb}{0.9, 0.9, 0.98}
\definecolor{indigo}{rgb}{0.29, 0.0, 0.51}

\usepackage[dvipsnames]{xcolor}

\usepackage{bbm}
\newcommand{\bo}{\mathbf{o}}
\newcommand{\bz}{\mathbf{z}}
\newcommand{\br}{\mathbf{r}}

\usepackage{xspace}

\newcommand{\silvio}[1]{{\color{Black}{#1}}}
\begin{document}

\title{Efficient Odd-One-Out Anomaly Detection}

\titlerunning{Efficient Odd-One-Out Anomaly Detection}

\author{Silvio Chito\and
Paolo Rabino\orcidlink{0009-0001-2843-0614} \and
Tatiana Tommasi\orcidlink{0000-0001-8229-7159}}

\authorrunning{S.~Chito et al.}

\institute{Politecnico di Torino, Italy 
\\
\email{\{name.surname\}@polito.it}}

\maketitle

\begin{abstract}
The recently introduced odd-one-out anomaly detection task involves identifying the odd-looking instances within a multi-object scene. This problem presents several challenges for modern deep learning models, demanding spatial reasoning across multiple views and relational reasoning to understand context and generalize across varying object categories and layouts. We argue that these challenges must be addressed with efficiency in mind. 
To this end, we propose a DINO-based model that reduces the number of parameters by one third and shortens training time by a factor of \silvio{three} compared to the current state-of-the-art, while maintaining competitive performance. Our experimental evaluation also introduces a Multimodal Large Language Model baseline, providing insights into its current limitations in structured visual reasoning tasks. \silvio{The project page can be found at \url{https://silviochito.github.io/EfficientOddOneOut/}}
\end{abstract}

\section{Introduction}
The concept of anomaly is both fundamental and inherently ambiguous.  
While in general anomalies signal critical deviations from expected patterns, their very definition is context-dependent as the singularity of a sample hinges on the framework in which it occurs. In the computer vision literature, anomaly detection and segmentation have been formalized mainly focusing on semantic novelty with the identification of previously unseen categories \cite{Ahmed_Courville_2020,chan2021segmentmeifyoucan}, or searching for defects as localized structural variations from standard reference data \cite{pmlr-v139-deecke21a,bogdoll2022anomaly,Li_2024_CVPR,Zhao_2023_CVPR}. However, the concept of anomaly is broader and recent works have started to investigate logical inconsistencies \cite{Bergmann2022,aaai.v38i8.28703} and to consider multi-modal frameworks including depth \cite{zavrtanik2024cheating,roth2022towards} and language \cite{winclip,gu2024anomalygpt}. Still, most of these works address settings with isolated objects, while realistic applications usually involve scenes with multiple objects in complex layouts and call for anomaly detection methods capable of handling clutter while dealing with multi-view analysis to elaborate on 3D object geometry and appearance. Moreover, every single scene may individually drive its own meaning of normalcy on the basis of cross-object comparisons and relative anomaly criteria. This requires models with an advanced level of relational reasoning to detect the \emph{odd-looking} instances while generalizing to scenes including objects (both categories and instances) not observed during training.  

To push research in the described challenging direction and closer to real world requirements, a new framework has been recently introduced which formalizes anomaly detection as an odd-one-out task \cite{bhunia2024odd}. Given multi-view images, this problem consists of identifying the samples in a scene that deviate from the others. It is inspired by standard tests in cognitive science, often used to assess intelligence and to provide valuable insights into an agent’s ability to manage high-level understanding of perceptual similarity. 

When designing new models to solve this problem, it is crucial to prioritize efficiency and avoid unnecessarily complex architectures. Indeed, such multi-object scenarios including outliers are common in industrial production, where minimizing inference time is critical, along with reducing re-training efforts by improving generalization. In this work we align with these objectives by introducing streamlined approaches that leverage the internal representations and extensive semantic knowledge of pre-trained large-scale models. Our contributions are summarized as follows:
\begin{itemize}[leftmargin=*]
\item we explore the use of Multimodal Large Language Models (MLLM) to tackle the anomaly detection task in the odd-one-out setting. We present a pipeline to define a baseline approach and assess its performance; 
\item building on previous work \cite{bhunia2024odd}, we show how the self-supervised representation extracted by DINOv2 can be used in a direct and effective manner by introducing an approach that reduces by one third the number of model parameters and cuts training time by a factor of \silvio{three}
while incorporating a module that enhances relational reasoning and advances state-of-the-art results.
\end{itemize}

\section{Related Works}
The elusive nature of the anomaly concept has fostered an extensive body of research, aiming to frame it in a wide range of settings accompanied by CNN and Transformer-based solutions. 
Given access to annotated outlier data, the problem of anomaly detection can be formalized as a supervised task \cite{pang2021anomreview}. However, anomalies are generally rare: despite some efforts to define image collections of industrial objects with labeled scratches, dents, or missing parts \cite{Bergmann2021}, their annotation is costly and anyway limited, leading to models that may not be proficiently used in practice. Indeed, most of the literature has focused on unsupervised approaches that have access only to normal samples during training, either in abundance or in limited quantity, with the latter setting known as few-shot anomaly detection \cite{xie2023pushing,lv2025oneforall}. Usually the problem is restricted to a binary task, defining separate models for each object category \cite{huang2022registration, FangFast}, but in the last years the literature has been moving towards a more natural unified multi-class framework \cite{Zhao_2023_CVPR,ding2022catching}. 
Over time, research has expanded beyond early approaches based on pixel reconstruction which assume that a network trained on normal images can accurately reconstruct anomaly-free regions while failing on anomalous areas 
\cite{ZAVRTANIK2021107706,inpaintingICIAP}. This strategy has been extended to feature reconstruction \cite{FangFast,patra2024revisiting} and knowledge distillation \cite{ST-bergman,9577330}, under the hypothesis that a student network trained to mimic a teacher exposed only to normal data will underperform on anomalies. 
More recent works leverage the rich representations 
of large-scale pre-trained vision models 
by computing feature distances \cite{padim,roth2022towards}, though these methods often involve computationally expensive memorization and matching steps. Another line of research revisits the supervised setting by synthesizing anomalies through noise injection, image mixing \cite{li2021cutpaste}, or generative models \cite{liu2023simplenet,strater2025generalad}. Finally, the rise of large-scale multimodal language-vision models has enabled zero-shot anomaly detection via tailored prompting and the use of textual descriptions \cite{gu2024anomalygpt,winclip}.

The overview provided above covers literature focusing on 2D data. Anomaly detection has been mainly studied on images despite real world applications clearly need 3D spatial reasoning. The MVTech3D dataset was one of the first proposed for 3D anomaly benchmarks \cite{visapp22} and has been followed by a few others with growing cardinality and resolution as well as different 3D formalization (point cloud, multi-view) \cite{Wang_2024_CVPR,real3dad}. Still, they all capture isolated instances and the proposed methods inherit standard strategies from the 2D literature. Only recently this niche has been extended by considering different settings such as semantic novelty detection to recognize unknown categories \cite{alliegro2022dos, rabino2024send}, conditional anomaly detection to compare a query image to a reference shape \cite{bhunia2024look3d}, and odd-one-out contextual anomaly detection involving multi-object scenes \cite{bhunia2024odd}.

The method introduced in \cite{bhunia2024odd} employs a 2D CNN architecture to extract visual features, which are then projected into 3D space to represent the observed instances. These representations are subsequently mapped back to the 2D domain to align with the internal feature space of DINOv2 \cite{oquab2024dinov}, enabling a distillation strategy that captures object geometry while injecting open-world knowledge. However, this two-step mapping introduces unnecessary complexity, which could be mitigated by directly extracting DINOv2 features from the multi-view input images. In the final stage, the method compares object-centric feature volumes by performing voxel-level part matching. We argue that this process can be streamlined and integrated with a strategy that explicitly encodes the relative distinctiveness of each instance with respect to the average normalcy observed in the scene. We detail our proposed approach in the following section.

\begin{figure}[t]
    \centering
    \includegraphics[width=0.95\linewidth]{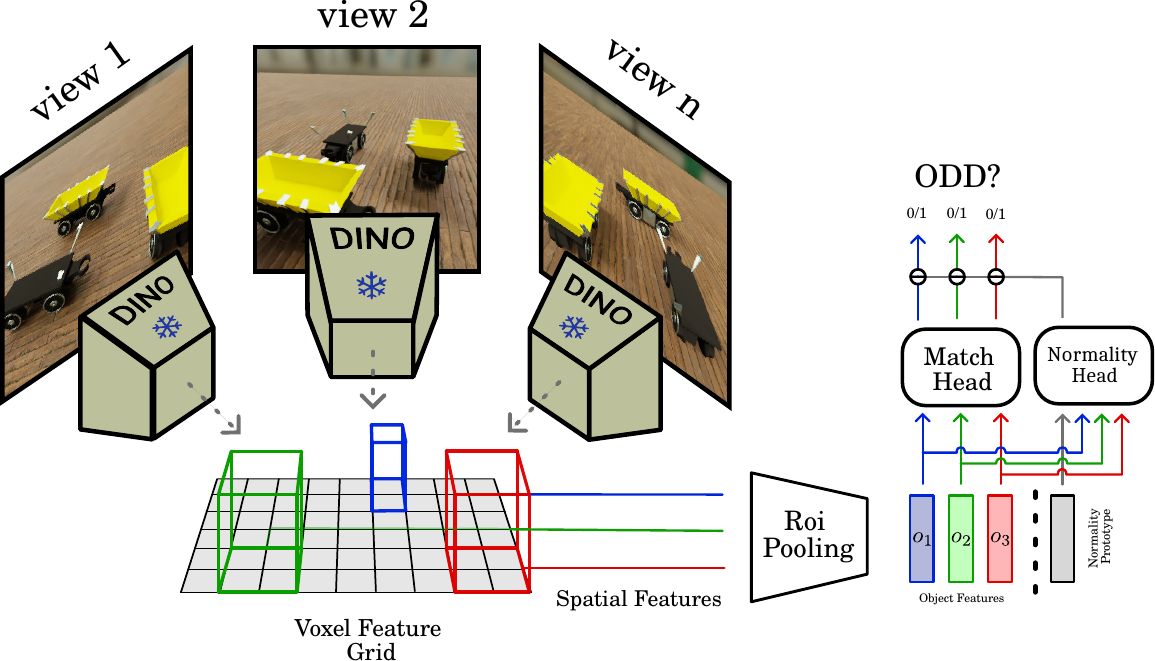}
    \caption{
    Schematic representation of our approach. It takes as input $M$ views of the scene, which are processed using the DINOv2 encoder. The resulting features are projected onto a voxel grid and associated with each of the $N=3$ objects. A subsequent pooling step yields per-object representations. These are further refined by the context and residual anomaly heads, which encode both object-to-object similarity and the relative deviations of each object from the scene-specific average normalcy. 
    }
    \label{fig:main_method} \vspace{-0.8cm}
\end{figure}

\section{Method}
We formalize the odd-one-out task by following \cite{bhunia2024odd} and considering a scene containing $N$ instances of an object category $o_{i=1}^N$. The goal is to identify which of these instances is odd, meaning that they present significant individual variations from the others that instead share a high visual similarity. We observe the scene from $M$ points of view corresponding to $\mathbf{I}_{t=1}^M$ RGB images with their camera projection matrices $\mathbf{P}_t$. Starting from these inputs and knowing the object locations in the scene, we need a model able to encode the appearance of separate instances and assign them a binary label $y_i\in \{0,1\}$,  where 1 identifies the anomaly. We target an efficient method that leverages the embedding space of pre-trained models and works with a minimal memory footprint (see Fig. \ref{fig:main_method}).

\textbf{Feature Encoder.} Several existing foundation models have shown impressive zero-shot performance in a variety of vision tasks thanks to their highly informative features that support strong generalization. In particular, DINOv2 \cite{oquab2024dinov} captures rich global and contextual cues, which are essential for accurately representing objects within complex 3D environments. We choose this model as a feature encoder for the multi-view scene images of our odd-one-out task. Thus, we obtain the features $\mathbf{F}_t=\texttt{DINOv2}(\mathbf{I}_t) \in \mathbb{R}^d$ for each input view, and project them into the 3D voxel space using the known camera parameters. The feature volumes are then averaged over all input views as in \cite{murez2020atlas, sun2021neucon} and fed to a 3D CNN-based network for refinement, obtaining the final representation $\mathbf{G}=\texttt{3DCNN}(aggr(\{\Pi_{proj}(\mathbf{F}_t,\mathbf{P}_t)\}_{t=1}^M)) \in \mathbb{R}^{d \times v_x \times v_y \times v_z}$.

\textbf{Context Match Head.}
From the obtained voxel feature grid, our goal is to extract the individual object instance features by using their 3D bounding boxes. 
However, working directly with 3D crops within the voxel grid poses several challenges: it is computationally expensive and can introduce artifacts due to the projection of high-dimensional DINOv2 features into 3D space, especially when object sizes vary. 
To mitigate these issues, we first normalize all object crops to a fixed $d\times8\times8\times8$ size using ROI pooling \cite{girshick2014rich}, then apply average pooling to squeeze their spatial representations into compact $d$ dimensional embeddings $\mathcal{O} = \{ \bo_1, \dots, \bo_N \}$, $\bo_i \in \mathbb{R}^d$. 
These instance representations are interpreted as tokens and provided as input to a standard transformer encoder of $L$ layers, allowing the model to reason about the relative appearance of objects in the same scene.
The output embeddings $\mathcal{Z} = \{ \bz_1, \dots, \bz_N \}$, $\bz_i \in \mathbb{R}^{q}$ reflect both the intrinsic features of individual objects and their relationships with others in the scene. 
We apply a simple linear layer to produce the final anomaly score, representing the probability $p$ that a given input is anomalous. The network is trained by minimizing the binary cross-entropy loss 
\begin{align} \label{BCE} \mathcal{L}_{\text{BCE}} = -\sum_{i=1}^{N} \left[ y_i \log(p_i) + (1 - y_i) \log(1 - p_i) \right]~.
\end{align}

\textbf{Residual Anomaly Head.} \label{sec:residual} 
Besides relying on pair-wise cross-instance matching, we can boost the identification of scene-specific anomalies by guiding the model to focus on how much an object deviates from the average normality. This deviation can be encoded into discriminative features via a residual anomaly module. 
We propose to devise it by adding to the network a 1-layer transformer encoder that takes as input the objects' representations $\mathcal{O}$ together with a learnable token $\mathbf{t}_0 \in \mathbb{R}^{d}$. The latter attends to all object embeddings, and the corresponding output  $\bz_0 \in \mathbb{R}^q$ is the only token retained for further processing. Specifically, it acts as a proxy for the scene-specific normality prototype and is supervised through a mean squared error loss 
\begin{equation} \label{mse_loss} 
\mathcal{L}_{\text{normality}} = \left\| \bar{\bz} - \bz_0\right\|_2^2~,
\end{equation}
where 
$\bar{\bz} = \frac{1}{|\{i \mid y_i = 0\}|} \sum_{i : y_i = 0} \bz_i$ is the average embedding of the ground-truth normal objects. 
Moreover, the residual anomaly module refines each object embedding via a projection head $f(\bz_i)$ (composed of linear layer, LayerNorm, and ReLU) to then calculate the difference between the obtained object representations and the learned centroid $\br_i = f(\bz_i)-\bz_0$. 

Finally, a linear layer elaborates on $\br_i$ to produce the anomaly score, representing the probability that a given input is anomalous. Thus, when the residual anomaly head is active, the network is trained by minimizing $\mathcal{L}_{\text{BCE}}+\mathcal{L}_{normality}$.

\section{The MLLM Baseline}
Recent advances in Multimodal Large Language Models (MLLMs) have demonstrated remarkable reasoning capabilities. By leveraging textual knowledge, these models significantly enrich visual understanding and enable a wide range of downstream tasks in a zero-shot setting (\ie without task-specific training), including anomaly detection \cite{gu2024anomalygpt,lv2025oneforall}. While MLLMs have proven effective in identifying anomalies based on the global appearance of objects, they still exhibit limitations in visual grounding \cite{elbanani2024probing}. Furthermore, their ability to reason about fine-grained anomalous details is currently an open question \cite{xu2025zeroshotanomalydetectionreasoning}, especially considering that most existing benchmarks are restricted to single-object scenes observed from a single viewpoint. 

\begin{figure}[t]
    \centering
    \includegraphics[width=0.95\linewidth]{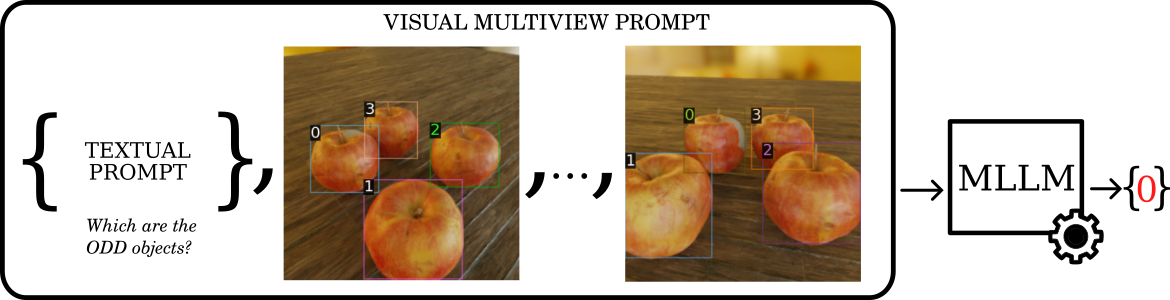}
    \caption{Representation of the Set-of-Mark \cite{yang2023setofmark} prompting strategy for the MLLM baseline. Note that the color of the bounding boxes and of the numerical indexes are randomized according to the SoM strategy.}
    \label{fig:MLLM} \vspace{-6mm}
\end{figure}

To explore the performance of MLLMs on the odd-one-out task, we present a tailored pipeline that leverages the Set-of-Mark (SoM) prompting method \cite{yang2023setofmark}. This strategy supports spatial understanding by detecting the objects in the scene and overlaying visual marks directly on the images, which are then provided as input to the MLLM. Specifically, we use gemini-flash 2.0 \cite{team2023gemini} and feed it the following prompt: \textit{Here is a list of images taken from multiple view-points of the same scene, each object is annotated with an index and the bounding box, one or more objects are different from the majority of all other objects, reply only with a list of indices of the odd objects}. The annotated images are concatenated to the prompt by following \cite{yang2023setofmark} as shown in Fig. \ref{fig:MLLM}. The model outputs the annotated index of the odd objects from which we can evaluate the prediction accuracy.  
\vspace{-2mm}

\section{Experimental Setup}

\textbf{Datasets.}
We adopt the same experimental setup of \cite{bhunia2024odd}, considering three tracks from two datasets containing objects with synthetically generated anomalies. \emph{Toys} is composed of 8K scenes with 3-6 objects from the 51 categories of the Toys4D dataset \cite{stojanov2021using} rendered in 20 views. The collection is divided into a training set with 5K scenes from 39 categories and two disjoint test sets: one (\emph{seen}) with 1K scenes from the seen categories but with unseen object instances, and the other (\emph{unseen}) containing 2K scenes from the remaining 12 novel categories. \emph{Parts} was defined from the ABC dataset \cite{koch2019abc} mostly containing mechanical parts shapes. The collection contains 15K scenes, each consisting of 3-12 objects rendered from 20 viewpoints, and is divided into train and test splits of respectively 12K and 3K scenes, with the latter including only \emph{unseen} shapes. As the datasets are provided\footnote{https://github.com/VICO-UoE/OddOneOutAD} with the anomalous objects always listed as first in the set, we ensured all labels were shuffled to avoid accidental information leakage and preserve the integrity of the evaluation. 

\smallskip\noindent\textbf{Baselines.} 
Our main baseline is the method that introduced the odd-one-out task in \cite{bhunia2024odd}, which we refer to as
\emph{OOO}. It included two multi-view object detection methods that we also adopt as reference: \emph{ImVoxelNet} \cite{rukhovich2022imvoxelnet} and \emph{Detr3D}. In addition, we use our newly defined zero-shot baseline that we indicate as \emph{MLLM}.

\smallskip\noindent\textbf{Metrics.} 
As in \cite{bhunia2024odd}, we evaluate the anomaly prediction performance via two metrics. The Area Under the ROC curve (\textit{AUC})  measures the ability of a model to discriminate odd from normal objects regardless of the chosen confidence threshold.  The \textit{Accuracy} considers an operating scenario with a threshold equal to 0.5.
In our analysis we feed the models with the ground truth location of the objects and focus only on the odd identification performance. This is justified by observations of the previous work \cite{bhunia2024odd} about the negligibility of the localization error due to the relative simplicity of the setting.
Note that for the zero-shot MLLM baseline we collect the Accuracy, but we do not have access to the confidence score so we omit the AUC.

\smallskip\noindent\textbf{Implementation details.}
All the models consider $M=5$ views as input, each with resolution $256\times256$. For our approach we use DINOv2 ViT-S/14 as feature encoder and a four-scale Encoder-Decoder-based 3D CNN as the 3D backbone.  While projecting the raw DINOv2 features (with DINOv2 frozen parameters) over the 3D voxel grid with dimension $96\times96\times16$, we use a bilinear interpolation for augmenting the height and width of the feature maps and then reduce channels from 384 to 32 with a $1\times1$ convolution resulting in a final voxel grid of dimension $32\times96\times96\times16$ (each voxel represents 4 cm). The object crops of dimensions $32\times8\times8\times8$ are flattened along the spatial dimensions through average pooling and projected into a 256 dimensional space via a linear layer. The Transformers employed by the \emph{Context Match Head} and \emph{Residual Anomaly Head} use a multi-head attention mechanism with 8 different 32-dimensional heads. 
We train our method on one NVIDIA V100 GPU for 50 epochs using a learning rate of $2\times10^{-4}$,
batch size 20, along with a cosine annealing scheduler and AdamW \cite{loshchilov2018decoupled} optimizer.

\begin{table}[t]
    \centering
    \resizebox{0.87\textwidth}{!}{
    \begin{tabular}{|@{~~}c@{~~}|@{~~}c@{~~}c@{~~}|@{~~}c@{~~}c@{~~}|@{~~}c@{~~}c@{~~}|}
    \hline
         \multirow{2}{*}{\textbf{Model}} & \multicolumn{2}{c|@{~~}}{\textbf{Toys Seen}} & \multicolumn{2}{c|@{~~}}{\textbf{Toys Unseen}} & \multicolumn{2}{c|}{\textbf{Parts}}\\
         & AUC & Accuracy & AUC & Accuracy & AUC & Accuracy\\
         \hline
         ImVoxelNet \cite{rukhovich2022imvoxelnet} & 78.13 & 65.55 & 73.19 & 60.12 & 72.80 & 64.34 \\
         DETR3D \cite{wang2021detrd} & 79.16 & 67.37 & 74.60 & 62.98 & 74.49 & 65.11 \\
         OOO \cite{bhunia2024odd} & \textbf{91.78} & 83.21 & \textbf{89.15} & \textbf{81.57} & 86.12 & 79.68 \\ 
         MLLM baseline & - & 52.23 & - & 53.35  & - & 60.73 \\

         \textbf{Ours}
         & 89.49 & \textbf{84.86} & 85.57 & 81.43 & \textbf{89.72} & \textbf{88.81}\\
         \hline
    \end{tabular}
    }
    \vspace{3mm}
    \caption{Main results for the three benchmark tracks. 
    Ours presents comparable performance to that of the leading competitor (OOO) on Toys and largely outperforms it on the challenging Parts dataset.}\vspace{-10mm}
    \label{tab:main}
\end{table}

\section {Results}
\textbf{Benchmark.} The results of our experimental analysis are presented in Table \ref{tab:main}. Our method performs similarly to OOO on Toys Seen and slightly worse on Toys Unseen, while showing a significant advantage over the leading competitor on Parts Unseen. We remark that the difference between these two datasets is in the geometric nature of the comprised objects: Toys includes free-form shapes with high inter-class variability, while Parts consists of mechanical components characterized by low semantic diversity and predominantly rigid angular geometries. 
Besides being more challenging due to the higher object count in the scenes and fine-grained shape differences, the limited categorical diversity of the latter dataset better reflects industrial quality inspection scenarios and aligns well with our approach, which is specifically designed to prioritize efficiency. 
We acknowledge that the OOO modules, which learn a task-specific 2D representation for each view, combine them in 3D and subsequently enhance them by passing through a differentiable rendering to distill DINOv2 features, offer greater flexibility in handling class variability.
However, this comes at a high cost in terms of model complexity\silvio{, including increased training time and a large number of parameters, due to the sparse voxel attention employed in the matching head.}

The MLLM baseline falls short on the Toys tracks, and on Parts it gets accuracy results close to those of the detection methods ImVoxelsNet and DETR3D. Overall, MLLM shows a bias towards normal data that might have been part of its large scale pre-training and, similarly to the detection methods, it recognizes large cracks and fractures but struggles when intra-group comparisons over multiple views are necessary. These limitations align with what was discussed in \cite{xu2025zeroshotanomalydetectionreasoning} and call for new tailored modules to guide generalist models when dealing with tasks requiring multi-view consistency.

\begin{figure}[tb]
    \centering
    \includegraphics[width=1.0\linewidth]{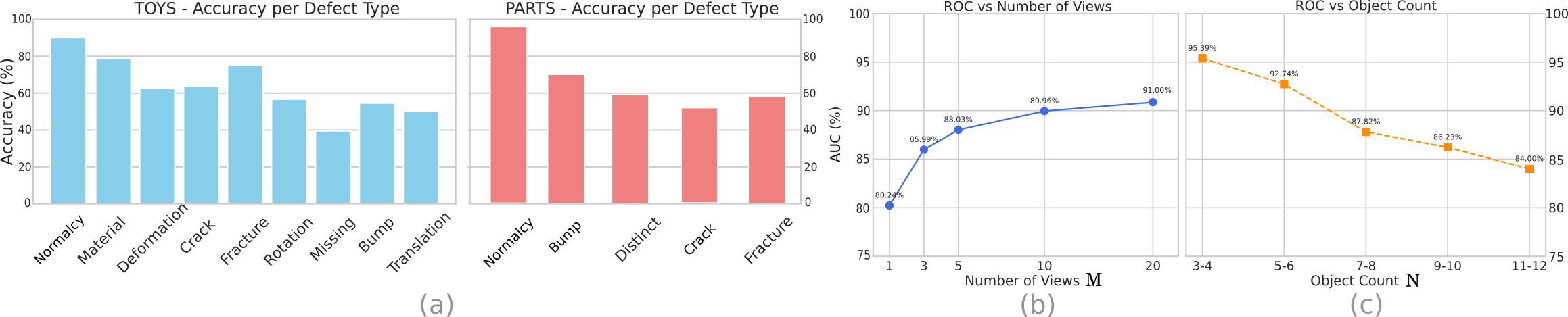} 
\caption{We report (a) accuracy across different anomaly types and normalcy data on Toys Unseen and Parts Unseen; AUC on Parts Unseen when changing (b) the number of views at inference time and (c) the number of objects in the scene.}
    \label{fig:anomaly_accuracy} \vspace{-1mm}
\end{figure}

\begin{figure}[t]
    \centering
    \includegraphics[width=0.9\linewidth]{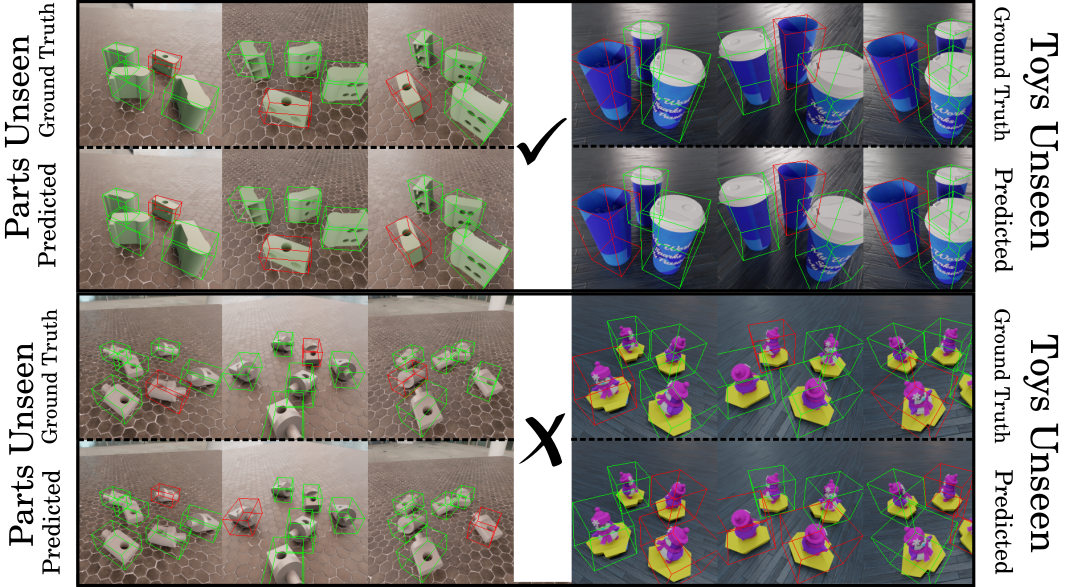}
    \caption{Qualitative results of our approach on Toys Unseen and Parts Unseen. We show three views for each scene. The green boxes indicate normal objects while the red ones indicate anomalous objects. The \cmark~ and \xmark~ indicate respectively correct and wrong predictions.}
    \label{fig:qualitative_res}
    \vspace{-4mm}
\end{figure}

\smallskip\noindent\textbf{Detailed analysis and Robustness.} We show a breakdown of the Accuracy of our approach across different anomaly types in Fig. \ref{fig:anomaly_accuracy} (a), together with success and failure qualitative results in Fig. \ref{fig:qualitative_res}. We can see from the breakdown that our model is particularly good at detecting material issues, bumps, and fractures that usually consist of darker regions of the object, while it struggles with 3D specific anomalies like missing parts, translations, and deformations that require a more global 3D reasoning. 
Furthermore, the right part of Fig. \ref{fig:anomaly_accuracy}  shows the AUC of our model when varying the number of input views during testing (b) and the performance when changing the object count (c). Both results are obtained on the Parts Unseen dataset and highlight the advantage provided by a growing amount of images as well as the relative robustness across scenes becoming progressively more difficult with a range of 4\% point variation around the main result (Accuracy 88.81 in Table \ref{tab:main}). 

\begin{table}[t]
    \centering
    \resizebox{0.9\textwidth}{!}{
    \begin{tabular}{|l@{~~}|@{~~}c@{~~}c@{~~}|@{~~}c@{~~}c@{~~}|@{~~}c@{~~}c@{~~}|@{~~}c@{~~}|@{~~}c@{~~}|}
    \hline
         \multirow{2}{*}{\textbf{Ours Head}} & \multicolumn{2}{c|@{~~}}{\textbf{Toys Seen~~}} & \multicolumn{2}{c|@{~~}}{\textbf{Toys Unseen~~ }} & \multicolumn{2}{c|@{~~}}{\textbf{Parts~~}} & \textbf{Memory} & \textbf{Inf. time}\\
         & AUC & Acc. & AUC & Acc. & AUC & Acc. & (GB) & (ms)\\
         \hline
         Sparse Voxel Attn. & 86.11 & 78.79 & 85.48 & 77.32 & 86.32 & 84.06 & 1.23 & 337\\
         Context & 89.18 & \textbf{85.42} & 85.09 & 81.27 & 89.14 & 88.65 & \textbf{0.36} & \textbf{271}\\
         Context + Residual  & \textbf{89.49} & 84.86 & \textbf{85.57} & \textbf{81.43} & \textbf{89.72} & \textbf{88.81} & 0.54 & 286\\
         \hline
    \end{tabular}
    }
    \vspace{0.3em}
    \caption{Ablation results across the three benchmark tracks for various versions of our model's head.}\vspace{-9mm}
    \label{tab:abl}
\end{table}

\smallskip\noindent\textbf{Ablations.} Among the architectural design choices of our method, there is the \emph{dense attention} on the pooled representations adopted within the Context Match Head. This is the standard choice for Transformers, but the original OOO approach exploited a different \emph{sparse voxel attention}. Building on voxel features certainly provides highly detailed information, but comes with the need for sparsity to avoid severe inefficiency. Thus, the cross-instance matching in OOO was performed considering only geometrically corresponding voxel locations. To explore the potential of the sparse voxel attention, we integrated it into our Context Match Head by replacing the original dense attention mechanism.  For this experiment we keep the same configuration proposed \cite{bhunia2024odd} by running our model for 50 epochs with batch size 4 and two different fixed learning rates for Toys ($2\times10^{-5}$) and Parts ($2\times10^{-6}$).

In Table \ref{tab:abl} we compare different versions of our model, also ablating the Residual Anomaly Head. From the obtained results we can conclude that the sparse voxel attention performs worse than the other versions on average. Moreover, it consumes significantly more memory and is considerably slower. Adding the Residual Anomaly Head slightly boosts the performance with a minimal change in memory and inference time. 
\silvio{The sparse version of the model also uses significantly more parameters (72M vs. 48M) and requires much longer training time, lasting about 6h on Toys and 22h on Parts (our models take around 2.5h on Toys and 5.5h on Parts).}

\section{Conclusions}
In this work we focused on the odd-one-out anomaly detection task that poses several challenges by moving anomaly detection closer to real world conditions. It involves multi-object scenes with scene-specific anomaly criteria that can only be inferred by cross-instance comparisons. This problem calls for models able to manage spatial reasoning to elaborate on multiple scene views, along with relational reasoning to interpret the context and generalize over different object categories and layouts. 
Considering the relevance of the task for industrial applications, efficiency is also a key requirement. By focusing on this aspect, we presented a model that, despite the significantly low capacity (number of parameters) and memory needs,
gets competitive performance compared to the state of the art. Several in-depth analyses confirmed the effectiveness of the adopted architectural choices as well as the overall model's robustness. 

To test the zero-shot effectiveness of Multimodal Large Language Models on the odd-one-out task, we also introduced a baseline that leverages the Set-of-Mark prompting method to support visual grounding. The low experimental results revealed the limitations of this approach, signaling the need for new solutions to leverage the extensive internal knowledge of generalist models when performing fine-grained comparisons in complex multi-view settings.

We believe that future solutions for the odd-one-out task should continue to build on the capabilities of pre-trained foundation models, with efficiency remaining a key objective. This would enable the transition from anomaly detection to full 3D anomaly localization, while allowing models to articulate the rationale behind the predictions through natural language. Furthermore, the task could be extended to encompass logical and functional aspects, bridging perception and robotics, and evolving into a true intelligence test for autonomous agents. 

\medskip
\noindent\textbf{Acknowledgments.} This study was carried out within the FAIR - Future Artificial Intelligence Research and received funding from the European Union Next-GenerationEU (PIANO NAZIONALE DI RIPRESA E RESILIENZA (PNRR) – MISSIONE 4 COMPONENTE 2, INVESTIMENTO 1.3 – D.D. 1555 11/10/2022, PE00000013). 
This manuscript reflects only the authors’ views and opinions, neither the European Union nor the European Commission can be considered responsible for them.

\vspace{-2mm}

\bibliographystyle{splncs04}
\bibliography{main}

\end{document}